\def\year{2018}\relax
\documentclass[letterpaper]{article} %DO NOT CHANGE THIS
\usepackage{aaai18}  %Required
\usepackage{times}  %Required
\usepackage{helvet}  %Required
\usepackage{courier}  %Required
\usepackage{url}  %Required
\usepackage{graphicx}  %Required
\usepackage{caption}
\usepackage{subcaption}
\usepackage{multirow}
\usepackage{amsmath}
\usepackage{array}
\usepackage[ruled,vlined,boxed,linesnumbered]{algorithm2e}

\begin{document}

\title{Recover Missing Sensor Data with Iterative Imputing Network}
\author{Jingguang Zhou, Zili Huang\\
	Shanghai Jiao Tong University\\
    \{wintersky, huangziliandy\}@sjtu.edu.cn
}
\maketitle
\begin{abstract}
	Sensor data has been playing an important role in machine learning tasks, complementary to the human-annotated data that is usually rather costly. However, due to systematic or accidental mis-operations, sensor data comes very often with a variety of missing values, resulting in considerable difficulties in the follow-up analysis and visualization. Previous work imputes the missing values by interpolating in the observational feature space, without consulting any latent (hidden) dynamics. In contrast, our model captures the latent complex temporal dynamics by summarizing each observation's context with a novel Iterative Imputing Network, thus significantly outperforms previous work on the benchmark Beijing air quality and meteorological dataset. Our model also yields consistent superiority over other methods in cases of different missing rates.   
\end{abstract}

\section{Introduction}
\noindent Big Data is indispensable for the development of machine learning \cite{manyika2011big}. 
Besides human-annotated data, geo-distributed sensors are great source for data collection, which benefits the development of machine learning methods in understanding the environmental dynamics. In a common sensing or crowd sensing campaign \cite{chong2003sensor}, sensors at different locations collect the environmental data during a time period.  However, most sensing campaigns suffer from systematic or accidental missing data mechanism, like broken sensors, communication errors and etc. Such unfortunate information loss throws importance upon imputing missing values in the sensor data. 

Sensor data recovery is a great challenge due to the remarkable portion of missing entries and their stochastic distribution. 
A handful of studies attempt to leverage the locality in the observational feature space 
via conventional methods like inverse distance weighting \cite{chen2012estimation} or ARMA \cite{valipour2013comparison} or nearest neighbors \cite{pan2010k}. 
Such methods yield unsatisfactory results as they fail to capture the latent, complex, and potentially higher-order temporal dynamics. 
In contrast, we aim to capture such dynamics in the latent (hidden) feature space by neural networks, which have proven rather effective in learning latent dynamics of time-series data~\cite{langkvist2014review,mei2017neuralhawkes}. 

However, common deep learning procedure cannot directly be used with incomplete training data. We develop a flexible scheme to deal with this----first initialize the entries using simple statistic estimates, and then update the estimated value via a novel multi-layer Iterative Imputing Network (IIN).
The core component of our Iterative Imputing Network is a multi-layer Long Short-Term Memory (LSTM) network, which consumes a sequence of time-stamped items and summarizes the representation and context information for each of them. An output layer is then stacked on this LSTM and projects the representation of any missing observation to a readable imputation. 
We propose to use two different versions of LSTM---the standard LSTM \cite{hochreiter1997long} for regularly sampled sensor data and Phase-LSTM~\cite{neil2016phased} for the irregularly sampled case.
We call it Imputing Network (IN).

Our novel Iterative Imputing Network (IIN) is a multi-level cascade of Imputing Networks (IN) that share the same set of weights, with the output imputation of any member IN block being fed into the higher-level one as input. 
It is thus mathematically equivalent to iteratively training the same Imputing Network with the same sequence, until the imputation accuracy achieves a satisfactory level.

Why is this important? Because by iteratively connecting (training) the network, the issue of data sparsity, known as a natural enemy of neural models, is well handled---the Imputation Network is able to gradually adapt itself by iteratively refining its missing value imputation on one single sequence sample.
Note that such data sparsity issue is especially troublesome in the task of missing data imputation, because 1) missing values naturally and effectively reduce data adequacy; 2) missing values sometimes are clustered together forming missing blocks, which are even more challenging to deal with.

Our Iterative Imputing Network has essential advantages over previous methods. 
First, our model summarizes higher-order temporal dynamics in the time-series, by representing each time-stamped observation with a deep neural network, while previous methods highly reply on locality in the observational feature space and could only summarize low-order (if more than one) temporal dependency in the time-series. 
Second, by representing the sequences in hidden space and iteratively refining the missing value imputation, our model better deals with missing blocks. Our model is consistent with the imputations within a missing block, and effectively adopt more information than the previous methods that skip missing values. 
According to these advantages, our model outperforms all previous methods on a hard benchmark Beijing air quality and meteorological dataset. Moreover, we demonstrate with our experiments that the superiority of our model is consistent with varying missing data rates.

In summary, our main contributions are as follows:
\begin{itemize}
	\item We propose a practical scheme for sensor data recovery, enabling the use of deep learning procedure by initializing missing entries via flexible methods.
	\item We design a novel Iterative Imputing Network (IIN), capturing the high-order latent temporal dynamics and iteratively refining the estimation of missing values.
	\item Our method significantly improves the state-of-the-art
	result on a hard benchmark, and shows robustness with varying missing rates.
\end{itemize}

\section{Related Work}
In the following, we review existing works related to our problem, including: (1) sensor data recovery; (2) deep learning for time series.  
\subsection{Sensing Data Recovery}
When wireless sensor network emerged, \cite{doherty2000algorithms} pointed out the importance of data recovery. Once all missing values have been imputed, the dataset can then be analyzed using standard techniques for complete data. 
\cite{yozgatligil2013comparison,lee2008missing} considered the temporal dependencies of sensing data series using statistical analysis like ARMA.
Some studies include the spatial cue into the missing data recovery. \cite{pan2010k} presented a K-nearest neighbor method for jointly spatial and temporal data imputation. \cite{yi2016st} considered spatial similarities together with their temporal similarities. 
Nevertheless, both of them only captured features on the surface, falling short in learning the internal dynamic of the temporal data.  \cite{gruenwald2010dems} applied tree-based data mining techniques to handle missing data on real-life and synthetic datasets.  \cite{lindstrom2014flexible} utilized matrix analysis, while \cite{sorjamaa2010improved} proposed a linear projection method called empirical orthogonal functions. However, their methods did not show enough robustness to the high rates and random distribution of missing data. Additionally, our approaches do not explicitly model the spatial similarities, since they can involve great uncertainties or noise. Instead, we feed all sensors' data into one network without seperating each sensor, enabling our network to benefit from shared trends and common property of different sensors.

\subsection{Deep Learning for Series}
Deep learning has been known for its great capability of learning data representations automatically, instead of using hand-craft features. Recurrent neural networks (RNN) and Long short-term memory (LSTM) networks \cite{gers2000applying,malhotra2015long} showed the efficacy for time series regression.
However, few studies dive deep in dealing with missing values. \cite{parveen2004speech} used binary indicators to handle missing values representing a pause between two sentences or a possibly interrupt for speech signals. Their missing values did not contain much information, which could cause much degradation of accuracy. They did not truly solve missing data recovery, but just ingore that. In contrast, our scheme aims to tackle the problem of missing data recovery, which has great significance due to the great amount but low quality of sensor data. Inspired by \cite{Dai2016Instance} which proposed a segmentation cascade model in computer vision, we design our multi-level cascade Iterative Imputing Network (IIN) to model the recovery of time series.

\section{Overview}
\begin{figure}[!h]
	\centering
	\begin{subfigure}{0.45\columnwidth}
		\centering
		\includegraphics[width=\linewidth]{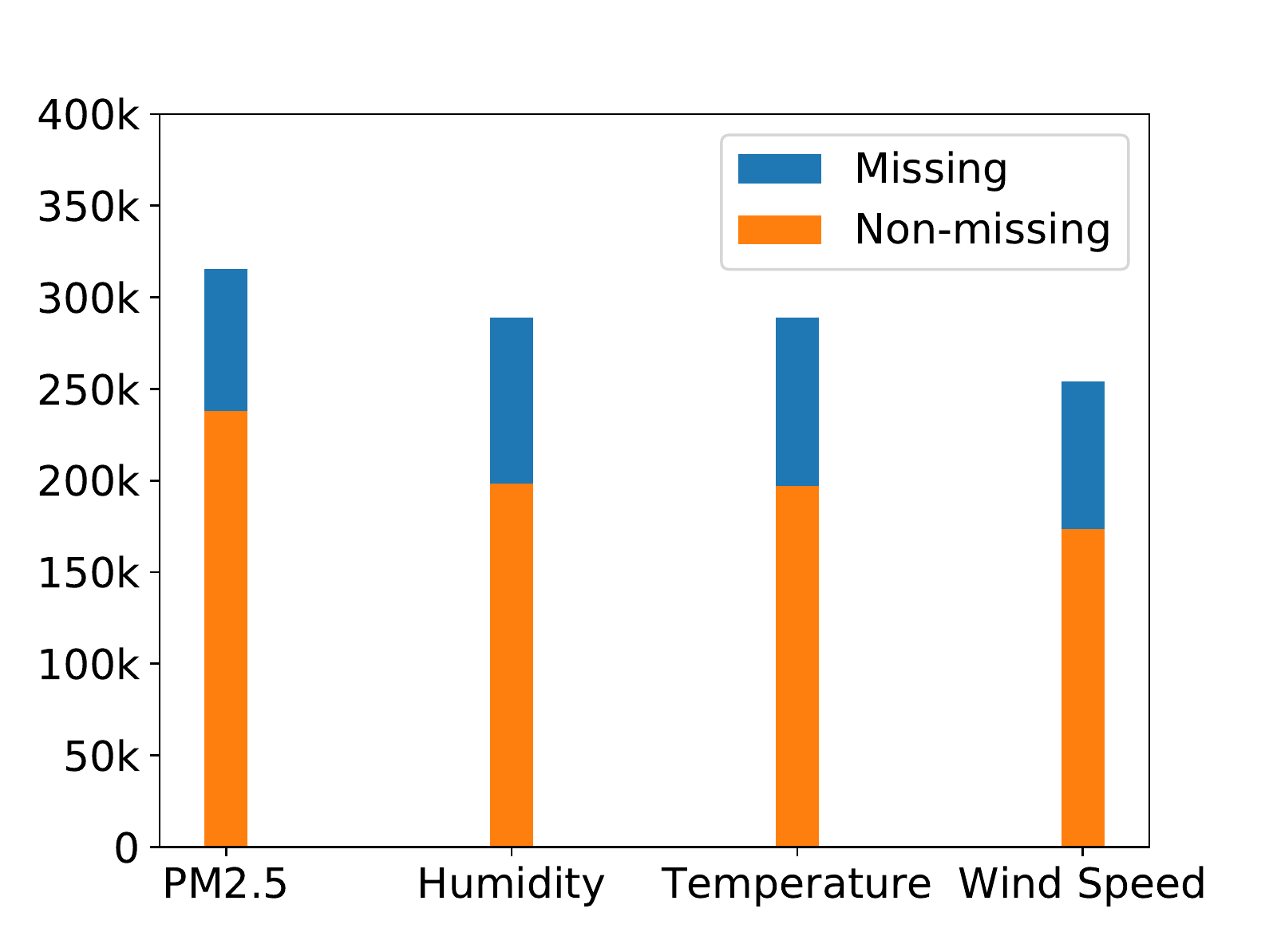}
		\caption{}\label{proportion}
	\end{subfigure}
	\begin{subfigure}{0.45\columnwidth}
		\centering
		\includegraphics[width=\linewidth]{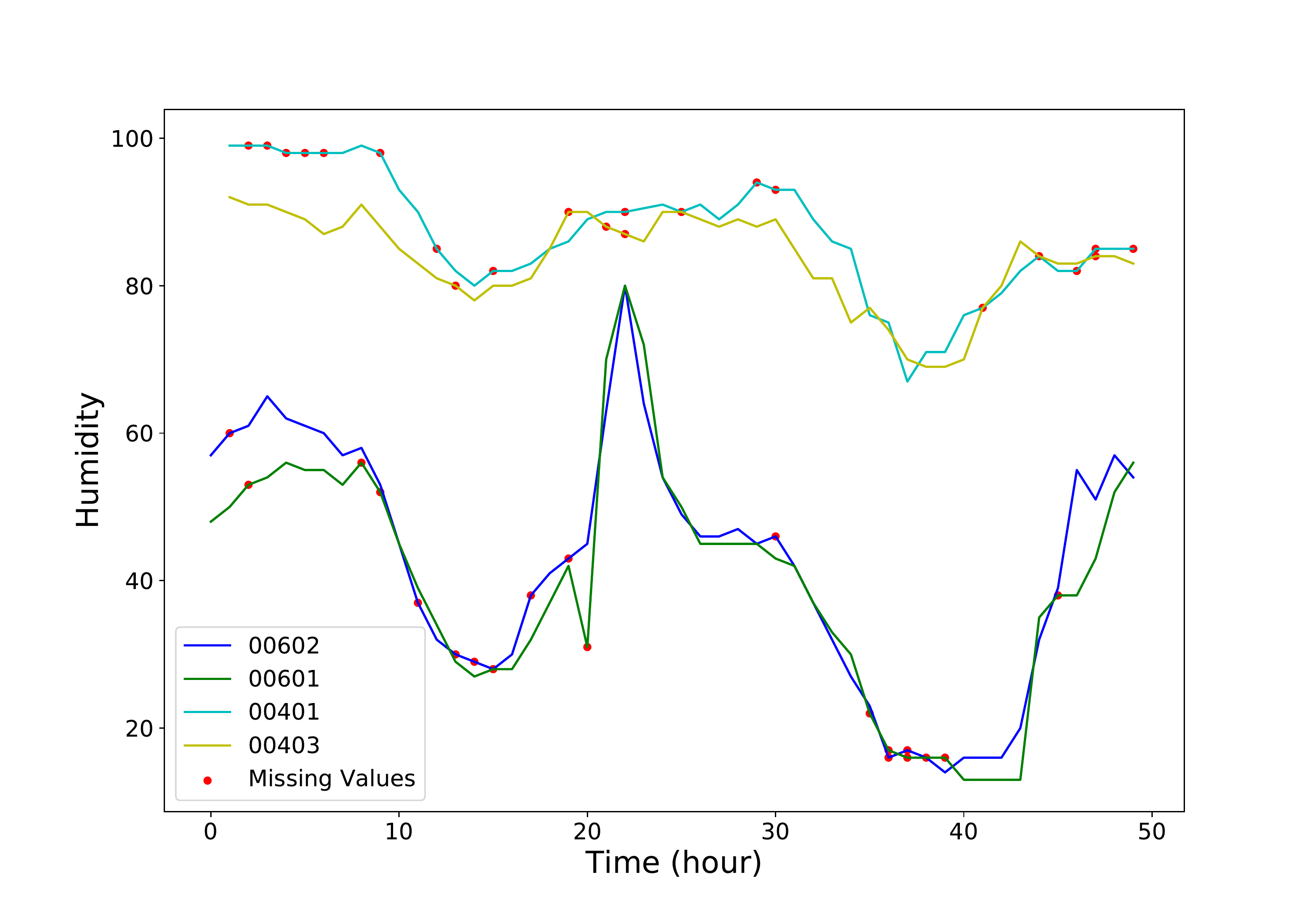}
		\caption{}\label{sensordata}
	\end{subfigure}
	\caption{(a) The proportion of missing data in four training datasets. (b) Humidity data collected by four sensors in two different locations. Red circle represents missing entries.}
\end{figure}

Sensor data usually suffer from missing values because of errors in data collection and transmission. The missing entries have large and random distribution, illustrated by Figure~\ref{proportion}.  It is hard for us to utilize the sensor data for subsequent analysis and visualization unless we tackle the prevalent innate missing entries in the dataset. 

To formulate the problem, we treat the series of data collected by one of the sensors as a sequence $\mathcal{X}$. Partial entries $\mathcal{X}_m=\{x_{m1}, x_{m2},..., x_{mk}\}$ are missing, while the other entries $\mathcal{X}_r = \mathcal{X}/\mathcal{X}_m$ are numerical values like temperatures and humidities. $\mathcal{X}_m$ are in an unknown random distribution and some of them locate in consecutive values, forming a block missing. Our goal is to learn from $\mathcal{X}/\mathcal{X}_m$ and get a better-estimated value for missing readings $\mathcal{X}_m$. Based on $\mathcal{X}$, we split the data into the training set $\mathcal{X}^{tr}$ and test set $\mathcal{X}^{te}$. For testing set $\mathcal{X}^{te}$, let $\mathcal{X}_r^{te}$ denote $\mathcal{X}^{te}/\mathcal{X}_m^{te}$. Then we extract a portion of non-missing entries denoted by $\mathcal{X}_{testGT}$ from $\mathcal{X}_r^{te}$ as ground truth, and learn how to recover from $\mathcal{X}_r^{te}/\mathcal{X}_{testGT}$. Our intuition is that first initialize the entries using simple methods with few costs, and then apply deep learning model to learn and give prediction based on  high-order latent temporal dynamics. The mean absolute error (MAE) and mean relative error (MRE) are used as metrics evaluating the quality of recovery data.

\section{The Model}
\subsection{Imputing Network}
\begin{figure}[htp]
	\centering
	\includegraphics[width=0.9\linewidth]{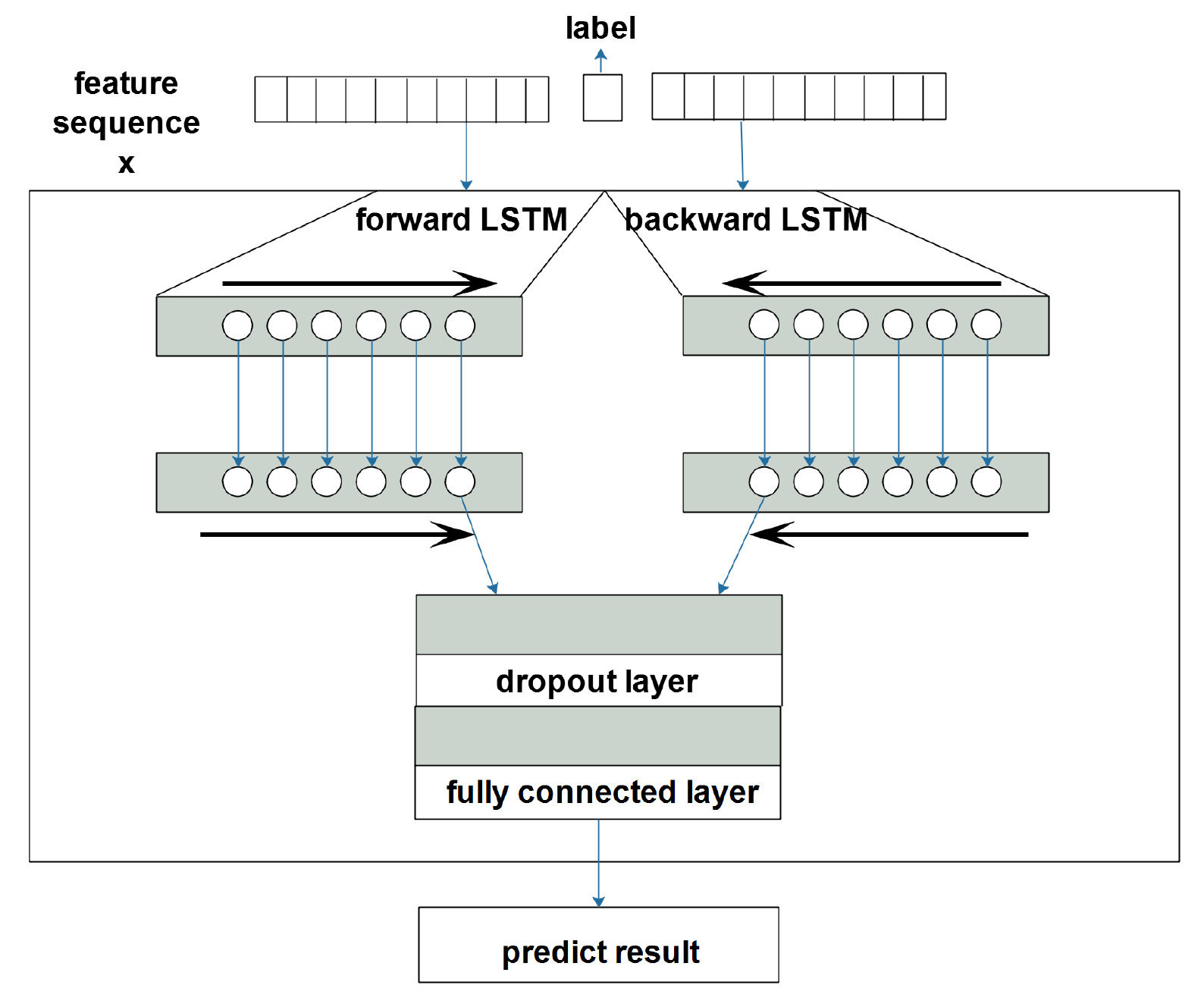}
	\caption{Architecture of Imputing Network.}
	\label{fig:model}
\end{figure}

Imputing Network, as the core component of our model, summarizes the context of each missing value by consuming its left and right neighboring observations or imputations with a forward and backward recurrent neural network (RNN) respectively, as shown in Figure~\ref{fig:model}. 
An output layer is then stacked upon the representations extracted by these RNNs and learns to impute the current missing value.

We choose Long Short-Term Memory (LSTM) networks as our RNNs, because they have proven very effective at sequential modeling---they are able to handle long-range dependency along sequences and to prevent the gradients from exploding or vanishing with the memory cell. 

Formally, our multi-layer LSTMs recurrently consume context of each to-be-imputed position $t$ in the sequence, as follow: 
\begin{subequations}
	\begin{align}
	h_s^{f} &= {LSTM}^{f}(h_{s-1}, x_s) \\
	h_u^{b} &= {LSTM}^{b}(h_{u+1}, x_u)
	\end{align}
\end{subequations}
where the superscripts $f$ and $b$ denote `forward' and `backward' respectively, $x$ is the (possibly imputed) observation at each position, and $s \leq t \leq u$. 
After $s$ and $u$ both reach $t$, the imputation $\hat{x}_t$ is computed by passing $h^{f}_{t}$ and $h^{b}_{t}$ through the output layer, as follow:
\begin{equation}
\hat{x_t} = OUTPUT(h^{f}_t, h^{b}_t)
\end{equation}. 
\begin{figure}[!h]
	\centering
	\includegraphics[width=0.9\linewidth]{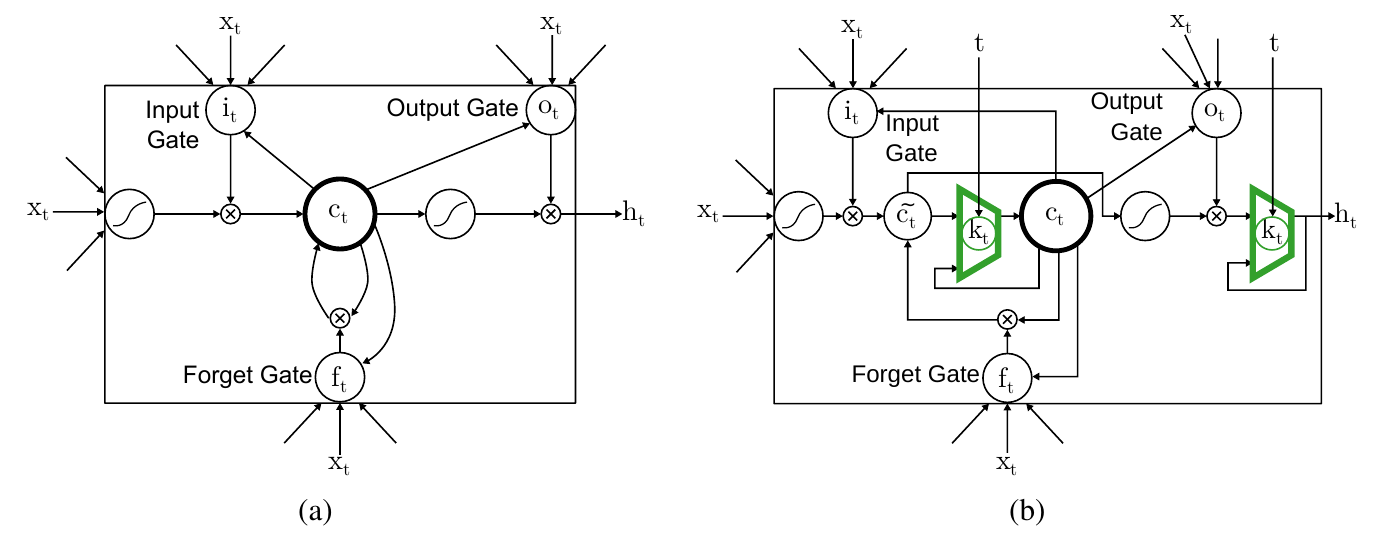}
	\caption{(a) standard LSTM: discrete-time; (b) phased LSTM: continuous-time.}
	\label{fig:lstms}
\end{figure}
In this paper, we adopt two versions of LSTMs, 
as shown in Figure~\ref{fig:lstms}.
The standard LSTM \cite{hochreiter1997long} is used to model regularly sampled sensor data. The update equations of standard LSTM is as follows.
\begin{subequations}
	\begin{align}
	i_t &= \sigma_i (x_t W_{xi} + h_{t-1}W_{hi} + w_{ci}\odot c_{t-1} + b_i) \\
	f_t &= \sigma_f (x_t W_{xf} + h_{t-1}W_{hf} + w_{cf}\odot c_{t-1} + b_f) \\
	c_t &= f_t \odot c_{t-1} + i_t \odot \sigma_c(x_tW_{xc} + h_{t-1}W_{hc} + b_c) \\
	o_t &= \sigma_o(x_t W_{xo} + h_{t-1}W_{ho} + w_{co} \odot c_t + b_o)\\
	h_t &= o_t \odot \sigma_h(c_t)
	\end{align}
\end{subequations}

Phase-LSTM~\cite{neil2016phased} incorporates time gate into LSTM cells in order to process irregularly sampled data, which is triggered by events generated in continuous-time~\footnote{For convenience, we call both two options LSTM and only differentiate them when it is needed.}. The update equations of Phase-LSTM are shown as follows:
\begin{subequations}
	\begin{align}
	\widetilde{c_j} &= f_j \odot c_{j-1} + i_j \odot \sigma_c(x_j W_{xc} + h_{j-1}W_{hc} + b_c) \\
	c_j &= k_j \odot \widetilde{c_j} + (1 - k_j) \odot c_{j-1}\\
	\widetilde{h_j} &= o_j \odot \sigma_h(\widetilde{c_j})\\
	h_j &= k_j \odot \widetilde{h_j} + (1 - k_j) \odot h_{j-1}
	\end{align}
\end{subequations}

\subsection{Iterative Imputing Network}
To learn refined imputation and deal with possible data sparsity caused by missing values, we propose a novel Iterative Imputing Network (IIN), which is a multi-level cascade of Imputing Networks that share the same set of weights, with the output imputation of any member IN block being fed into the higher-level one as input.
This is important (as we can show shortly in experiments), because such design enables the Imputing Network to gradually adapt itself by iteratively refining its missing value imputation on one single sequence sample! 
This benefits the model learning by 1) jointly utilizing the information from not only visible observations but also previously imputed missing values and 2) well handling (potential) data sparsity caused by missing data. 

\begin{figure}[!h]
	\centering
	\includegraphics[width=1.0\linewidth]{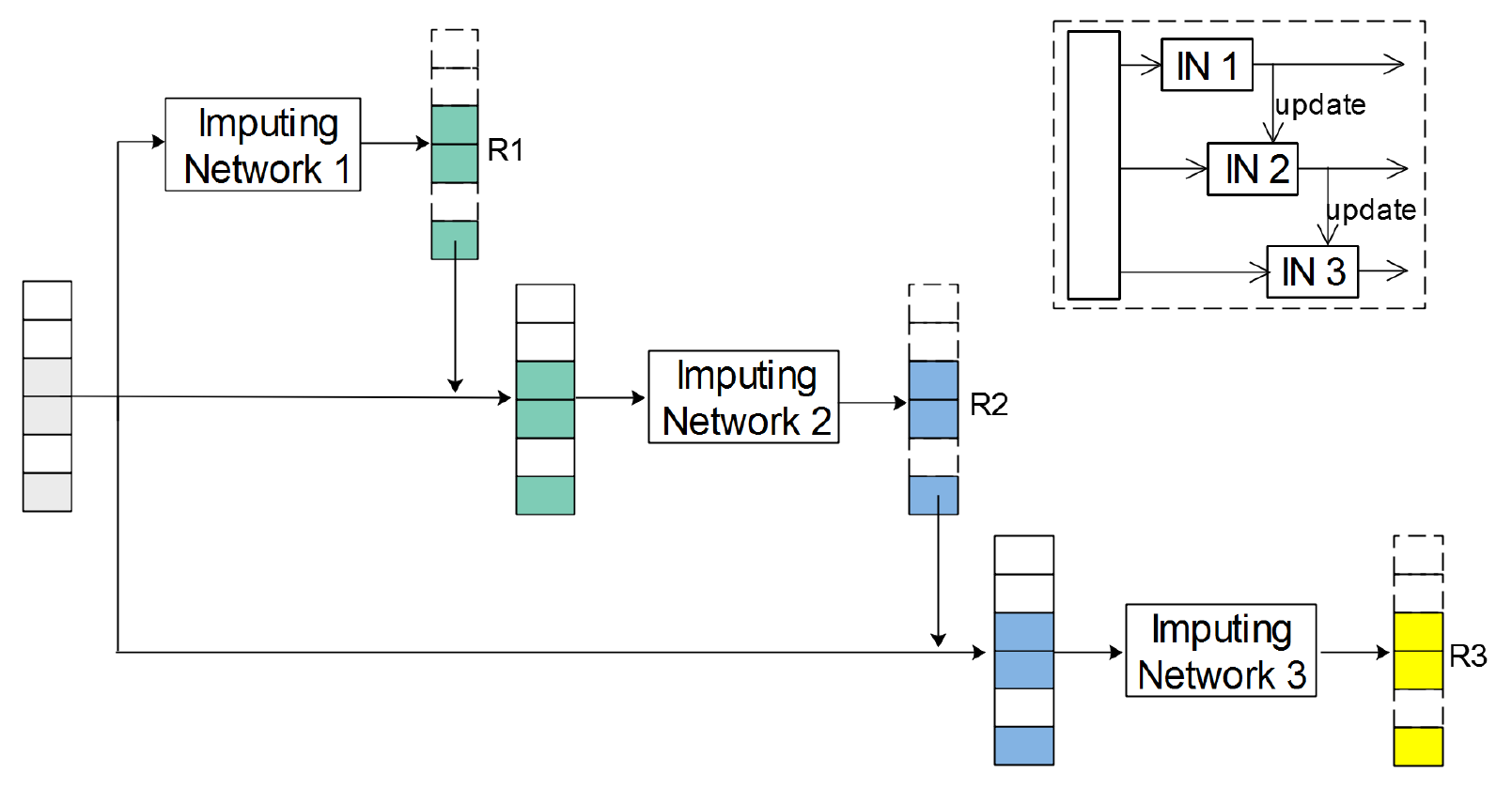}
	\caption{A three-stage cascade. On stage 2, series in which missing entries are updated by Imputing Network 1 are used as input to Imputing Network 2.}
	\label{fig:model2}
\end{figure}

Figure \ref{fig:model2} illustrates the INN architecture. Note that this design is mathematically equivalent to iteratively training the same Imputing Network with the same sequence, until the imputation accuracy achieves a satisfactory level. Therefore, we naturally have two options for training the model---training the Iterative Imputing Model as a whole by gradient-based methods or iteratively training the same Imputing Network. 
As the former option is straightforward, we only elaborate the latter one in this section. 

Our iterative recipe to train the model is illustrated in Algorithm~1.
\begin{algorithm}
	\caption{Iterative Training Recipe.}
	\KwIn{Origin Data Series $M$; Iterations $iter\_num$}
	\KwOut{Final Recovery Data Matrix }
	Initialize  series $T_0$ from $M$ with statistical methods;\\
	Fill the remaining missing entries in $T_0$ with nearest data records in time domain;\\
	$i \leftarrow 0$;\\
	\Repeat{$i \geq iter\_num$}{
		Extract valid feature-label pairs from $T_i$;\\
		Train Imputing Network $model_i$;\\
		Predict the missing values in $T_i$ using $model_i$;\\
		$T_{i+1} \leftarrow $ update the missing entries in $T_i$;\\
		$i \leftarrow i+1$;\\
	}
	return series $T_{i+1}$;
	\label{scheme} 
\end{algorithm}
This algorithm is essentially analogous to expectation-maximization (EM) algorithm: using currently-estimated model weights to impute the missing values (expectation) and then updating these weights by maximizing the log-likelihood given the observational data (maximization)~\cite{allison2002missing}.

\section{Experimental Analysis}
In this section, we evaluate the effectiveness of our model on the benchmark Beijing air quality and meteorological dataset~\cite{zheng2013uair}, elaborate the experimental details, and analyze the results.

\subsection{Dataset Preparation}
We conduct experiments on the Beijing air quality and meteorological dataset.
The geo-distributed air quality data and meteorological data was recorded every hour. 
We select the subset of PM2.5 from the air quality dataset and select the subsets of temperature (TEMP), humidity (HUM) and wind speed (WS) from the meteorological dataset.
Figure~\ref{sensordata} shows humidity data series collected by $4$ sensors in two different places. Sensor 00601 and 00602 are close to each other, while Sensor 00401 and 00403 are in the adjacent areas. 
Looking from the time dimension, the sensing data do not show an obvious periodicity. The sensing data may be affected by sparse asynchronous streams of events, which makes the learning and understanding of the internal data pattern a very challenging task. Worse still, there are noticeable missing entries in sensing dataset because of collecting errors and network errors, illustrated by Figure~\ref{proportion}.

Besides the original missing entries in the sensing dataset, we need to prepare our training set and testing set by setting aside some entries as ground-truth. 
For PM2.5 dataset, we apply the method in \cite{yi2016st} to generate missing values. First, we record the positions of all missing values in each month's data. Then we manually remove the values on the same position in the next month. (For instance, if the entry for a sensor at 2014-05-04 14:00:00 is missing, then we drop out the value of this sensor at 2014-06-04 14:00:00). For the other three datasets(temperature, humidity and wind speed), we randomly set aside $20\%$ of the total non-missing values as the ground-truth of missing entries. In our experiment, we use the sensor recordings in the $3$, $6$, $9$ and $12$ month as testing set and the rest as training set.

\subsection{Anchor Selection}

\begin{figure}[htp]
	\centering
	\includegraphics[width=0.9\linewidth]{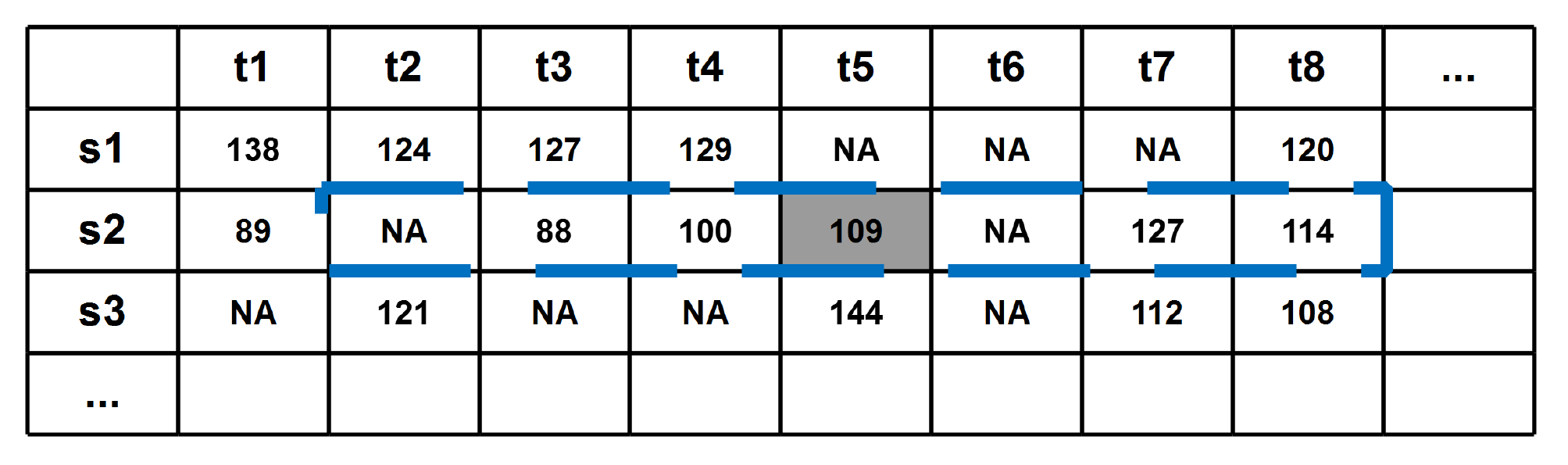}
	\caption{Anchor (blue) w.r.t. an entry (gray).}
	\label{anchor}
\end{figure}

A common practice in sensor data recovery is to operate within a sliding window that is centered at the to-be-imputed entry. Such a sliding window is usually called the {\em anchor}~\cite{ren2015faster} of this entry~\footnote{ An anchor is associated with a window size, which could be large in order to cover several sensing cycles.}~\footnote{Anchors can be extended to multi-scale, i.e. combining anchors with multiple window sizes, or covering multiple sensors at the same time, but here we discuss the basic anchor for a single series.}. An example anchor of size $7$ in the dataset is shown in Figure~\ref{anchor}. As we can see, the to-be-imputed entry is neighbored by its left (previous) and right (subsequent) context. The special value NA denotes the missing observations.

We aim to train our model only with anchors that carry adequate information about the dynamics. Therefore, we define each anchor to be valid for use, only if more than $50\%$ of the entries are observable (i.e. not missing). 
As with how to fill the missing blanks during training, we use the imputed values estimated by our model (with recently updated weights), which will be shown more effective than other common practices shortly in the results section. 

\begin{figure*}[t]
	\centering
	\begin{subfigure}{0.95\columnwidth}
		\centering
		\includegraphics[width=\linewidth]{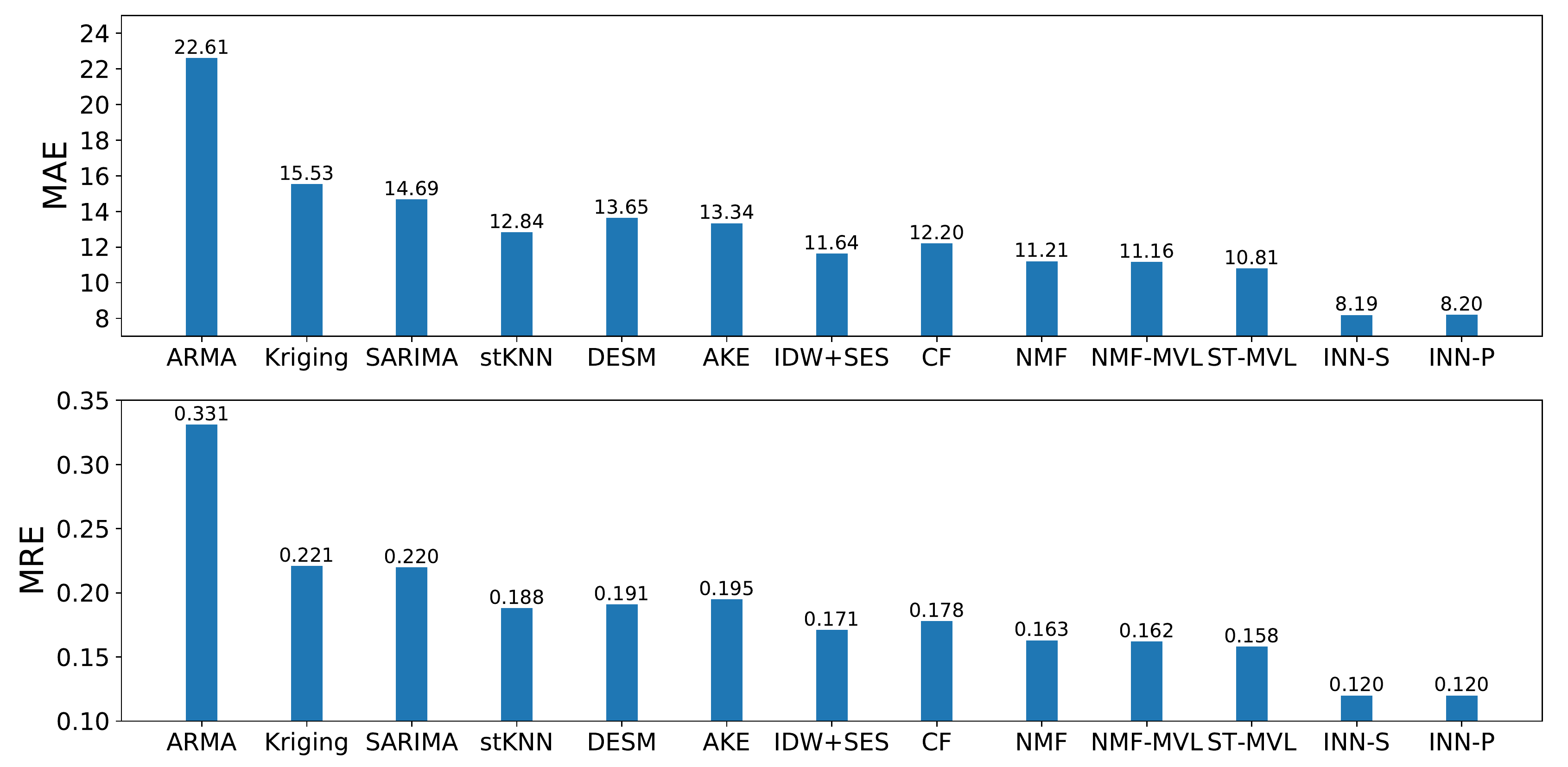}
		\caption{}\label{general}
	\end{subfigure}
	\begin{subfigure}{0.95\columnwidth}
		\centering
		\includegraphics[width=\linewidth]{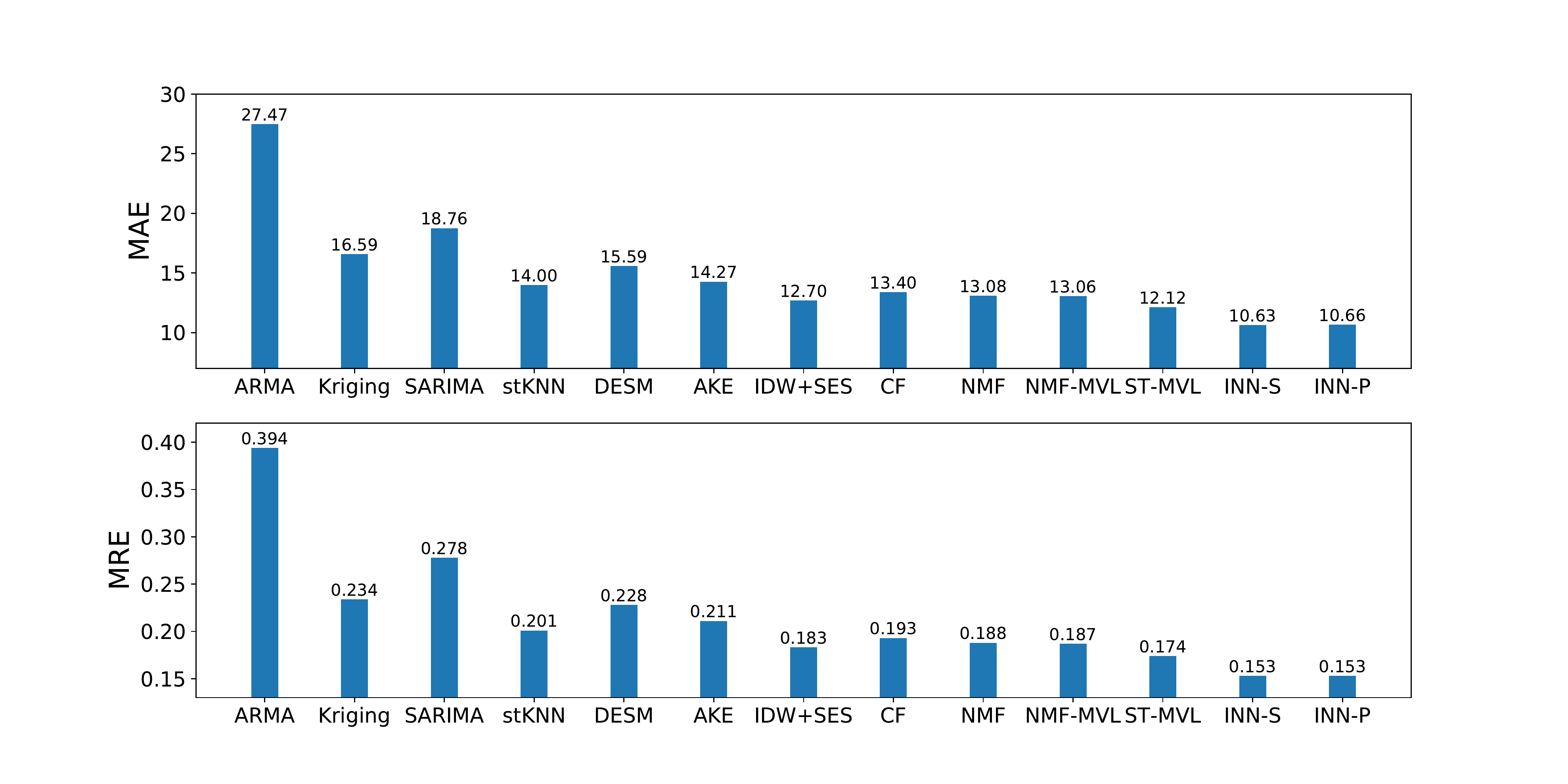}
		\caption{}\label{overall}
	\end{subfigure}
	\caption{Recovery error of different methods in two scenarios (a) general missing; (b) overall missing. }
	\label{different methods}
\end{figure*}

\subsection{Implementation Details}
In our Imputing Network, we use a two-layer forward and backward LSTM to model the latent patterns of series. The first layer takes a sequence with a half of the window size as input, then outputs hidden units each of which is of $50$ dimensions. The second layer will output the last $100$-dimensional unit. We concatenate the outputs of forward and backward LSTM and then use dropout at a rate of $0.3$. We set aside $1/10$ of the data in training set as our validation set and maintain the models which show best performance on the validation set. 

To train the Imputation Network, we use mean absolute error (MAE) as our loss function and we apply Nesterov-accelerated adaptive moment estimation (Nadam) algorithm \cite{sutskever2013importance,dozat2016incorporating} to optimize our neural network. Nadam incorporates Nesterov Momentum into Adam, making it consistently outperform Adam and RMSProp. For the LSTM cells, we initialize kernel weight matrix in a glorot uniform distribution, the recurrent kernel weight matrix in an orthogonal distribution, and set the bias to be zeros. 

To implement IIN, a multi-level cascade of Imputing Networks, we maintain the index of the missing entries. 
We iteratively update the missing entries and pass the output into Imputing Network for two or three times. To update the missing entries, we maintain their indices. Cascade IIN will refine the data recovery as the number of iteration increases.

The entire process can also be seen as the adaption of EM iterative strategy to deep learning.
At the first round, the large ratio of missing reading in a sensing dataset makes it unable to apply deep learning methods. Therefore, we actually use some combinations of statistical values of the data, which can be viewed as an unsupervised feature extraction. This is similar to E step. After we impute the missing entries from observed data, we get "artificially" intact data and pass them to the LSTM-based IIN networks. The Imputing Network will learn the dynamics of the sensor data and give the most probable outputs for missing entries. This is similar to M step. After the first round, we could omit the E step----use the network outputs in the previous round as input data and do M step directly.

\subsection{Recovery Accuracy and Error}

We measure the performance by Mean Absolute Error (MAE) and Mean Relative Error (MRE),
$$MAE=\frac{\sum_{i=1}^{N}|x_{t_i}-\hat{x}_{t_i}|}{N},MRE=\frac{\sum_{i=1}^{N}|x_{t_i}-\hat{x}_{t_i}|}{\sum_{i=1}^{N} x_{t_i}}$$
where $\hat{x}_{t_i}$ and $x_{t_i}$ is the estimated and ground-truth value respectively at time $t_i$ (with index $i$), and $N$ is the total number of observations.

Missing values occur in sensing dataset in a stochastical way. 
To eliminate the dominance of extreme cases and corner cases on MAE and MRE, we compute the mean error in a general scenario, which we denote as general missing. Similar to \cite{yi2016st}, for the general missing we do not consider the spatial missing block (the missing values that records of all sensors are simultaneously absent) and the temporal missing block (the records of a sensor are missing in a certain length of time window, $11$ in our experiments). For fair comparisons, we compute MAE and MRE for all missing entries, which scenario we denote as overall missing.

We compare our method with 10 baselines, including ARMA \cite{valipour2013comparison}, stKNN \cite{pan2010k}, Kriging \cite{wu2013spatial}, SARIMA \cite{yozgatligil2013comparison}, DESM \cite{gruenwald2010dems}, AKE \cite{pan2010k}, IDW+SES \cite{gardner1985exponential}, CF \cite{sarwar2001item-based,su2009a}, NMF \cite{lee2001algorithms,lindstrom2014a}, ST-MVL \cite{yi2016st}. Prior best results on Beijing air quality and meteorological dataset are achieved by ST-MVL, which use statistical methods to extract four features from the observed feature space. 

We evaluate two types of IIN, based on standard LSTM and phased LSTM respectively. We denote them as IIN-S and IIN-P. 
Table \ref{different methods} shows the comparison of different missing
scenarios between different methods~\footnote{Results of different methods come from \cite{yi2016st}.}. As we can see, IIN outperform other baselines significantly in two missing scenarios. ST-MVL, the prior best work, improves $0.3$ on general missing than before. In contrast, IIN's MAE improves $2.6$ on the results of ST-MVL. Moreover, IIN predict estimation for all spatial blocks and temporal blocks, while prior methods skip some extremely hard missing entries when computing the accuracy. In spite of this, IIN　 still outperforms significantly the prior methods. 
\begin{figure}[htp]
	\centering
	\includegraphics[width=1.0\linewidth]{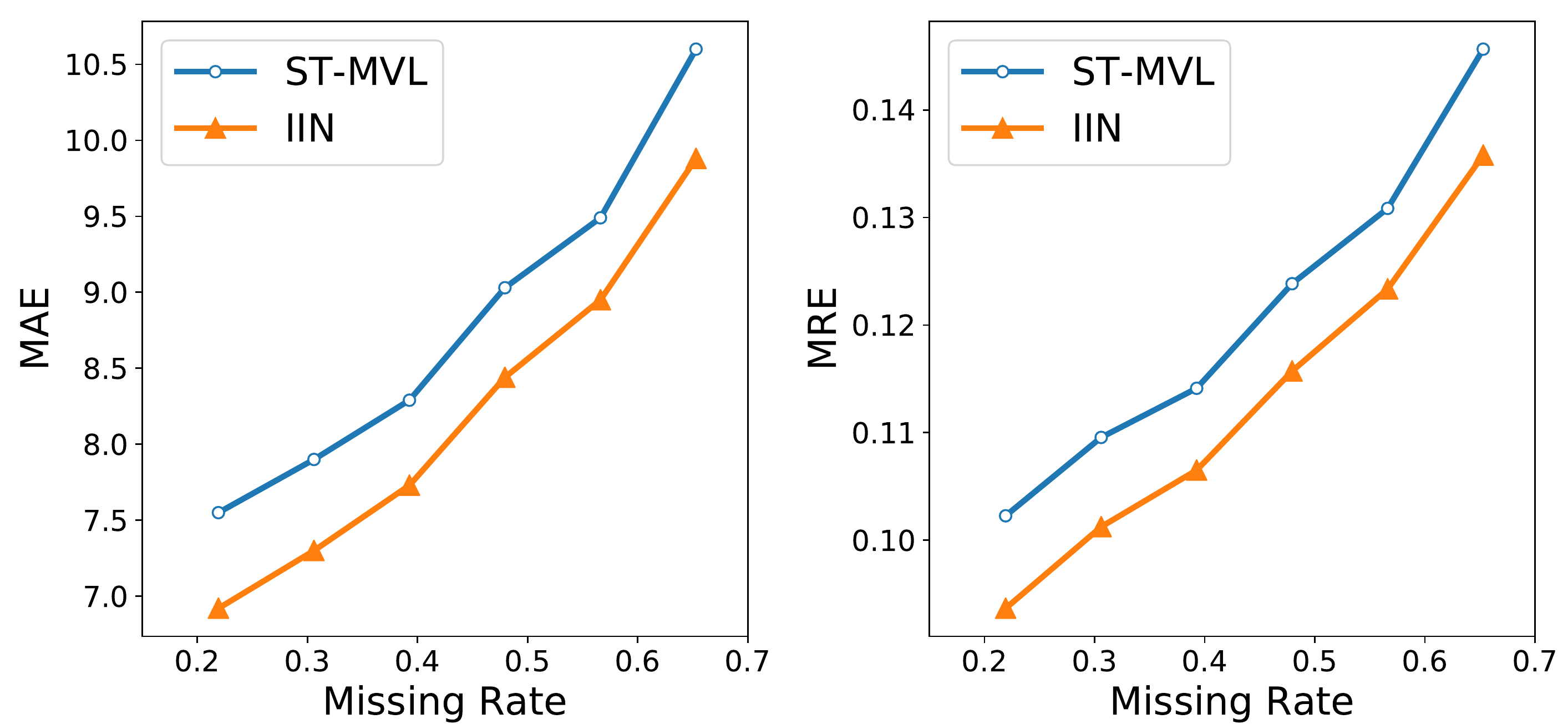}
	\caption{Impact of missing rates over the recovery error.}
	\label{miss_rate}
\end{figure}
\subsection{Performance with Different Missing Rates}
We consider the impact of different missing rates over our training process. The miss rate refers to the ratio of the number of missing entries over the total entries. Higher missing rate bring severe bias to data recovery. Here we prepare our dataset by randomly dropping out the data records with different missing rates.
Illustrated by Figure~\ref{miss_rate}, Our methods are always better than ST-MVL.   

\subsection{Missing Value Initialization Analysis}
To kick off the training of our model, we fill the missing entries with appropriate initialization. Our schemes are flexible in different initialization methods, and then refine the estimation using deep learning approaches. 
We initialize the missing entries using the method proposed by~\cite{yi2016st}, which regresses each missing entry initialization on four commonly used corpus-level statistics. 

\begin{figure}[htp]
	\centering
	\includegraphics[width=0.95\linewidth]{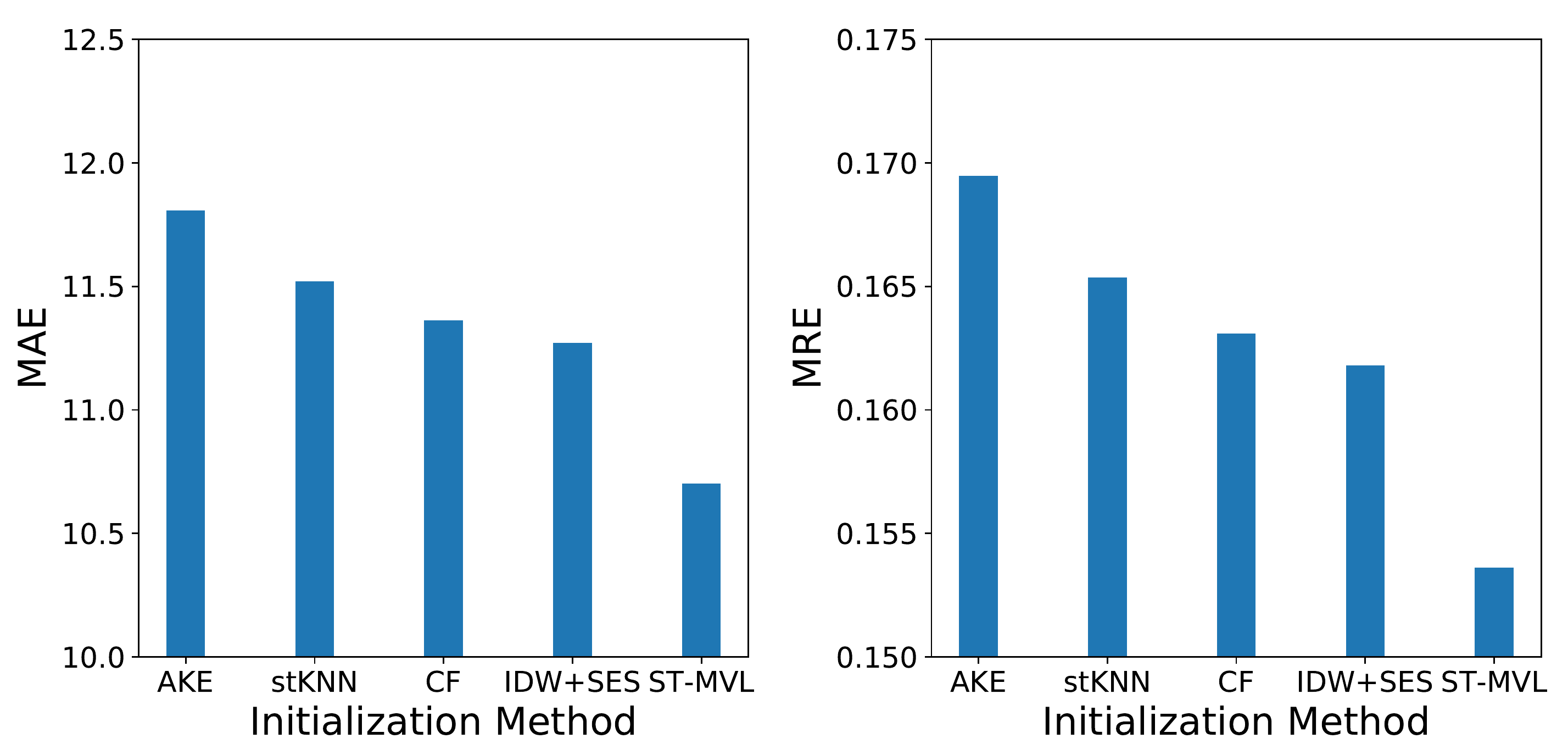}
	\caption{Impact of initialization method based on PM2.5.}
	\label{Init_method}
\end{figure}

In Figure~\ref{Init_method}, we also evaluate the impacts of other initialization methods on our recovery accuracy. We compute the recovery error after passing the initialized data into Imputing Network. 
ST-MVL is the best initialization method, probably because it combines four statistical factors. Whatever initialization methods are used, IIN can always help refine the estimation of the results. (For MAE, AKE: $14.27$ to $11.77$; CF: $13.40$ to $11.41$; ST-MVL: $12.12$ to $10.66$.) We should note that even if we initialize with AKE, IIN can further help to reduces the MAE to $11.77$, lower than ST-MVL $12.12$ without IIN.

\subsection{Separate or Mix Different Sensors?}
In our recovery scheme, we utilize data from all sensors in order for jointly training of our neural networks. However, it seems natural to separately train and predict for the data of different sensors, in order to avoid their data patterns interfering with each other. We experimented in this way, applying our schemes on separate sensors and taking an average of their error. We denote this separate version as IIN(sep). The results are illustrated as Table~\ref{Table:four datasets, single}. 
IIN shows better performance on all datasets than the state-of-art methods ST-MVL. Nevertheless, IIN(sep) is not always better than ST-MVL. IIN(sep) shows an edge over ST-MVL on PM2.5 dataset, but fails on temperature dataset. 

Therefore,  mixing sensor data does not introduce much noise. Instead we argue that this enables our network to benefit from shared trends or common properties of different sensors.
Some geo-distributed sensor data have strong correlations, which benefits the recovery process on the missing data. For example, if we need to impute the missing values from an anchor of sensor 1, we find its neighboring sensor 2 has complete data of its anchor in the same period. Then the anchor of sensor 2 will help if we pass it into IIN, which gives prediction based on common data patterns in the latent feature space.

\begin{table}[t]
	\centering
	\caption{Results on different datasets.}
	\begin{tabular}{|r|r|p{0.8cm}<{\centering}|p{0.8cm}<{\centering}|p{0.8cm}<{\centering}|p{0.8cm}<{\centering}|}
		\cline{1-6}
		\multicolumn{2}{|c|}{Dataset}&PM2.5&TEMP&HUM & WS
		\\\cline{1-6}
		{\multirow{2}*{ST-MVL}}&MAE&12.12&0.68&3.37&1.89\\
		\cline{2-6}
		%&MRE&0.1505&0.0459&0.0591&0.2985
		&MRE&0.1740&0.0459&0.0591&0.2985
		\\\cline{1-6}
		
		{\multirow{2}*{IIN(Sep)}}&MAE&10.78&0.74&3.10&1.88\\
		\cline{2-6}
		&MRE&0.1558&0.0496&0.0544&0.2958
		\\\cline{1-6}
		
		{\multirow{2}*{IIN}}&MAE&10.63&0.63&2.90&1.87\\
		\cline{2-6}
		&MRE&0.1531&0.0422&0.0509&0.2953
		\\\cline{1-6}
	\end{tabular}
	\label{Table:four datasets, single}
\end{table}

\section{Conclusion}
Besides the human-annotated data that is usually rather costly, sensor data has been for a long time playing an important role in machine learning tasks. However, systematic or accidental mis-operations often result in a variety of missing data, which significantly adds up the noise in the collected dataset. While previous work only imputes the missing values by interpolating in the observational feature space, we aim to model the latent (hidden) temporal dynamics by summarizing each observation's context with a novel Iterative Imputing Network. Our model significantly outperforms previous work on the benchmark Beijing air quality and meteorological dataset, and also yields consistent superiority over other methods in cases of different missing rates.

\section{Acknowledgments}
We sincerely thank Hongyuan Mei for many helpful discussions and comments on the manuscript. We also thank Xinsong Zhang and Weijia Jia for their introduction of the dataset and their encouragement.

\bibliographystyle{aaai}
\bibliography{bare_conf}

\end{document}